\documentclass[runningheads]{llncs}
\usepackage{graphicx}
\usepackage[]{todonotes}
\usepackage{float}
\usepackage{hyperref}
\usepackage{multirow}

\begin{document}
\title{Training Multimodal Systems for Classification with Multiple Objectives}
%
%
\author{Jason Armitage\inst{1} \and
Shramana Thakur\inst{1} \and
Rishi Tripathi\inst{1} \and
Jens Lehmann\inst{1, 2} \and 
Maria Maleshkova\inst{1}}
\authorrunning{Jason Armitage et al.}
%
\institute{University of Bonn, Germany \and
Fraunhofer IAIS, Dresden, Germany}
\maketitle              
\begin{abstract}
We learn about the world from a diverse range of sensory information. Automated systems lack this ability as investigation has centred on processing information presented in a single form. Adapting architectures to learn from multiple modalities creates the potential to learn rich representations of the world - but current multimodal systems only deliver marginal improvements on unimodal approaches. Neural networks learn sampling noise during training with the result that performance on unseen data is degraded. This research introduces a second objective over the multimodal fusion process learned with variational inference. Regularisation methods are implemented in the inner training loop to control variance and the modular structure stabilises performance as additional neurons are added to layers. This framework is evaluated on a multilabel classification task with textual and visual inputs to demonstrate the potential for multiple objectives and probabilistic methods to lower variance and improve generalisation.

\end{abstract}
\makeatletter
\def\blfootnote{\xdef\@thefnmark{}\@footnotetext}
\makeatother
\section{Introduction}\blfootnote{Copyright © 2020 for this paper by its authors. Use permitted under Creative Commons License Attribution 4.0 International (CC BY 4.0).}
Human experience of the world is rooted in our ability to process and integrate information present in diverse perceptual modalities~\cite{sepulcre_stepwise_2012}. Multimodal approaches to machine learning are motivated by this ability and aim to develop rich representations that combine information from multiple sources~\cite{bruni_multimodal_2013}. Consider a machine learning system that models events in the world by processing inputs from online media sources. Representations of these events take the form of text, images, video, and audio. Systems that are able to process signals from a range of these inputs learn models that are more complete descriptions of the represented events with resulting benefits for inference and predictions~\cite{baltrusaitis_multimodal_2017}. Researchers have proposed related classifiers for performing event detection~\cite{petkos_social_2012}, source prediction~\cite{ramisa_multimodal_2017}, and activity recognition ~\cite{yang_deep_2017}.

Multimodal machine learning presents a suite of methods for leveraging diverse data - but the development of systems that generalise to unseen samples leads to challenges arising both in practice and from the underlying theory of machine learning. Limited data resources are the most pressing concern in the first category. Data acquisition for multimodal systems is complicated by the requirement for combinations of samples in each input modality~\cite{amato_ai_2019}. In the absence of large-scale data, neural networks learn sampling noise in the training data and report low scores on unseen samples~\cite{srivastava_dropout_nodate}. Additional modalities also inflate parameter counts with the outcome that multimodal systems report high accuracy during training and low accuracy at test time~\cite{wang_what_2019}. Additional hidden layers can boost performance during training but introduce the requirement to prevent the interaction of parameters across the model from slowing convergence~\cite{ioffe_batch_2015}.

Multimodal fusion combines representations from constituent modalities into a single embedding. In recent years, the deployment of neural networks to generate fused embeddings has resulted in state-of-the-art performance on classification tasks of textual and visual samples. In theory, multimodal fusion methods capture information present in the input representations and produce outputs with complimentary information. Comparison with unimodal classifiers demonstrates that the introduction of additional modalities yields only modest performance gains~\cite{wang_what_2019}. In addition to overfitting, limitations on available data are acute when tasks require images or video. 

\textbf{Main contributions.} We propose and build a novel approach to multimodal classification that introduces a second objective to learn fused embeddings trained with variational inference. To our knowledge, this use of multiple objectives - where one function is learned with a method from inverse probability - is unique in the research on multimodal representation learning. We go on to show that the range of methods for calibrating parameter updates developed within the latter approach offsets the overfitting associated with multimodal fusion. The benefits of these proposals are demonstrated empirically by adapting an existing end-to-end architecture to perform multilabel classification on a dataset of 25k samples of paired images and text. $F$-scores on classifying unseen samples provide measurement of the contribution from introducing a second probabilistic objective and related regularisation methods to multimodal classification tasks.  

\textbf{Structure of the analysis.} We start with an outline of the use case identified for multimodal representation learning followed by a detailed specification of our proposed framework and related methods. The evaluation section presents topline results for the system and an ablation on regularisation methods in variational inference. Section 4 highlights the existing research informing our work and we conclude by summarising the main findings.  

\section{Use Case}

Learning on multiple modalities presents opportunities to generate rich representations for enhancing performance on existing tasks and enables new applications~\cite{baltrusaitis_multimodal_2017}. This section introduces a form of classification task where samples are presented both in natural language and images. Systems that learn on these modalities are applied to a range of use cases related to archived and online media~\cite{ramisa_multimodal_2017,petkos_social_2012,baltrusaitis_multimodal_2017}.

\subsection{Multilabel Classification on Multiple Modalities}
Label prediction underlies approaches to information retrieval~\cite{chen_machine_1995} and classification~\cite{aggarwal_survey_2012} - and enables the downstream tasks of document retrieval~\cite{hinton_reducing_2006}, textual and visual entailment~\cite{marelli_sick_nodate,xie_visual_2019}, and fact validation~\cite{lehmann_defacto_2012}. Assigning samples to a potential subset of multiple labels also characterises challenges in unimodal~\cite{nam_learning_nodate} and multimodal~\cite{kiela_efficient_2018} real-world applications.

Multilabel classification on image and text inputs forms a benchmark task in the research on bimodal learning~\cite{kiela_efficient_2018,pang_multimodal_2011} and is also used here to assess our proposed approach to multimodal fusion. In creating the MM-IMDb dataset for movie genre classification, the authors were addressing a shortfall in training data to conduct multimodal classification~\cite{arevalo_gated_2017}. The task in MM-IMDb is an instance of multilabel prediction over multiple modalities where titles have an average of 2.48 classes and the system undertakes a series of independent classifications. Metrics are computed by comparing outputs $Y$ with target labels from the set $D$. The authors propose an architecture (referenced below as the \emph{GMU baseline}) with gates to control information learned from modalities to perform classification. We build a version of this system in PyTorch and include results as a benchmark (see Figure \hyperref[fig1]{2} and Table \hyperref[tab1]{1}). 

\subsection{Dataset}
The MM-IMDb dataset constitutes samples for 25,959 movies assigned to one or more of 23 genre classes. Inputs processed in predicting class labels for each sample are a text summary averaging 92.5 words and an image poster. An additional 50 metadata fields of structured text are excluded from the multilabel prediction task to enable assessment of systems on natural language and images. 

\begin{figure}[h!]
\includegraphics[width=\textwidth]{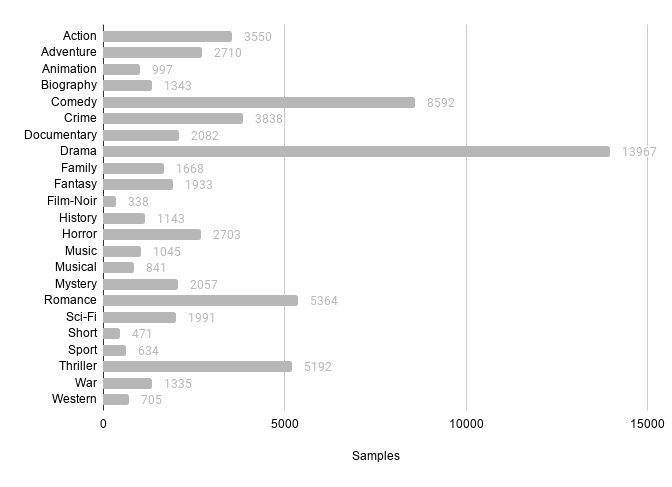}
\caption{Number of samples attributed to each genre in the MM-IMDb dataset.} \label{fig0}
\end{figure}

\renewcommand{\thefootnote}{$\star$} 
\renewcommand{\thefootnote}{*}
Systems are evaluated on a processed version of the dataset available from the authors' institution \footnote{\url{http://lisi1.unal.edu.co/mmimdb/multimodal_imdb.hdf5}}. We extracted four columns from MM-IMDb: FileID, Genres (in one-hot encoded format), VGG16 image embeddings, word2vec text embeddings.  Text embeddings are $300-dimension$ word2vec representations and image embeddings are $4096-dimension$ representations of features extracted by Arevalo \textit{et al.}~\cite{arevalo_gated_2017}. Image embeddings, text embeddings, and one-hot-encoded true labels are stored as separate tensors ahead of training. Training and cross-validation are performed with 70\% of the data and systems are evaluated on the remaining 30\% of samples.  

\newpage
\subsection{Variance in Multimodal Classification Systems}
\begin{figure}
\includegraphics[width=\textwidth]{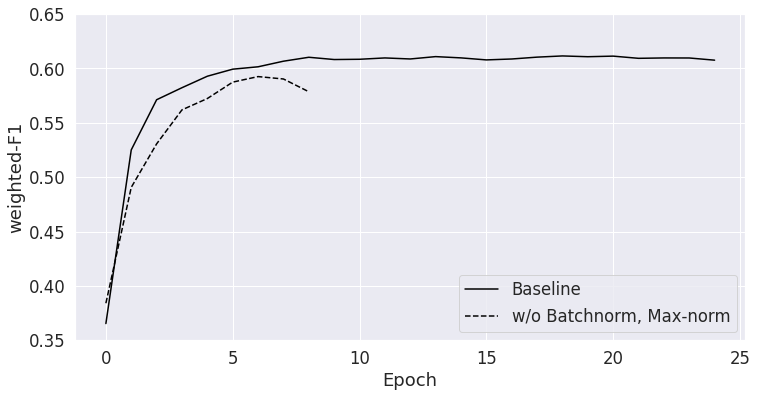}
\caption{Performance on the validation set of MM-IMDb by $weighted-F1$ for the GMU baseline system and a version excluding regularisation methods. Curves are plotted to maximum score +1 epoch.} \label{fig1}
\end{figure}

High variance is a core challenge to system performance at test time in multimodal classification tasks~\cite{radu_multimodal_2018}. Arevalo \textit{et al.}~\cite{arevalo_gated_2017} note the improvements that regularisation methods contribute to the architecture proposed for conducting classification on the MM-IMDb dataset. As a starting point, we examine the effects of excluding batchnorm~\cite{ioffe_batch_2015} and constraining weight updates to an upper bound (max-norm)~\cite{srivastava_dropout_nodate} on system performance. Validation accuracy curves in Figure \hyperref[fig1]{2} run to one epoch after the maximum $weighted-F1$ score (see Section 4) observed for different versions of the baseline system. The negative impact of variance is visible after only a few epochs when these methods are excluded from the system.   

\newpage
\section{Approach}
We have examined the importance of regularisation in the use case above and continue with our proposal to provide additional controls for calibrating updates to a set of parameters $\Theta$. In this section, we introduce a classification framework with multimodal fusion comprised of two modules trained with separate loss functions and a single computation of gradients w.r.t. inputs. This framework is the basis for our investigation into mitigating variance with the aim of improving classification performance over multiple modalities and is referred to as \emph{PM+MO} below. 

\subsection{Classification Framework with Multimodal Fusion}

A multilabel classification framework with bimodal inputs is a function $h(x)$ that takes pairs of samples $(x_i^t,x_i^i) =((x_1^t,x_1^i),(x_2^t,x_2^i),...(x_n^t,x_n^i))$ where $t$ and $i$ are text and image representations respectively. The resulting multimodal representations are mapped to a subset of labels $S \subseteq D$ or $d_i$ and each output is classified $y=(y_1,y_2,...y_n)\in\{0,1\}.$ In the proposed framework, the two objectives in $h(f_1(x),f_2(z))$ are computed sequentially with a single step of gradient computations. Module $A$ learns the function $f_1(x)$ for multimodal fusion with a variational inference framework and \hyperref[sec:multimodal_fusion]{ELBO} as the objective. Module $C$ conducts multilabel classification $f_2(z)$ on the outputs of $A$ by optimising a standard loss described directly \hyperref[bce]{below}.

Three wide hidden layers are the basis of the fused embedding module $A$. As with Arevalo \textit{et al.}~\cite{arevalo_gated_2017}, input embeddings $v^t$ and $v^i$ are assigned to linear functions and hyperbolic tangent $tanh(u_i^l)$ activations where $u$ is the hyperbolic angle. In our proposal, weights and biases of hidden layers $[w_{i,j}^{I,l} , b_i^l]$ are random variables drawn from a Laplace $L$ distribution $\theta_j^l\sim L(\mu_j,\sigma_j)$ with mean $\mu_j$ and variance $\sigma_j$ optimised during training. Outputs from the unimodal embedding layers are fused using concatenation (1) and mixing (2) operations:
\begin{equation}
v_{j}^{cat}=[v_{j}^{t,o},v_{j}^{i,o}]
\end{equation}
\begin{equation}
z_j =v_{j}^{cat} * v_{j}^{i,o} + (1-v_{j}^{cat})*v_{j}^{t,o}
\end{equation}
Loss on $A$ is computed with stochastic variational inference using the variants of ELBO detailed in Section 3.2. Multimodal embeddings $v^m$ form the inputs for the classifier module $C$. We align with Arevalo \textit{et al.}~\cite{arevalo_gated_2017} in implementing a multilayer perceptron with maxout activation $max_j(w_{i,j}^{I,l}x , b_i^l)$ as proposed by Goodfellow \textit{et al.}~\cite{goodfellow_maxout_2013}. A wide hidden layer receives $v^m$ and the $max$ of parameters for $v_[m,o]$ are taken during activation. \label{bce}Binary cross-entropy combines sigmoid activation with cross-entropy loss to assign a probability for each class to outputs $(y_1,y_2,...y_n)\in\{0,1\}$ and summing the results $ $$\sum_{c=1}^{M}y_{j,c},log(p_{j,c}).$ 

Regularisation methods recommended by Arevalo \textit{et al.}~\cite{arevalo_gated_2017} - and retained in our framework - consist of batchnorm to learn $\gamma z+ \beta$ for $\mu[z]$ and $[z]$ on batches $\beta$, max-norm to constrain weight updates $||w_j^l||$, and dropout with maxout activation in $C$. Both modules are optimised with variants of the Adam algorithm incorporating regularisation. Adam with gradient clipping in $A$ - as implemented in Pyro PPL~\cite{bingham_pyro_2018} - is run on each parameter during steps of variational inference. In the case of $C$, AdamW was used in place of Adam after initial testing. This algorithm acts on Loschilov and Hutter’s proposal to replace L2 regularisation with decoupled weight decay when Adam is the optimiser~\cite{loshchilov_decoupled_2019}. 

\subsection{Multimodal Fusion with Variational Inference}
\label{sec:multimodal_fusion}
Multimodal embeddings are learned by inferring $[w_{i,j}^{I,l} , b_i^l]$ for the layers of $A$ using variational inference. In each case, we assume that the posterior $p(\theta_j^l|X,Y)$ is drawn from a family of Laplace distributions $Q$. Computing the integrals for $p$ is intractable and so we infer an approximate candidate from $Q$ by selecting $q_i$ with the lowest KL divergence from $p$~\cite{wingate_automated_2013}.The posterior expression is reduced to $p(\theta_j^l|x)$ and defined as:
\begin{equation}
p(\theta_j^l|x) =\frac{p(\theta_j^l, x)}{p(x)}.
\end{equation}
Minimising the KL divergence
\begin{equation}
KL(q(\theta_j^l)||p(\theta_j^l|x))
\end{equation}
is equivalent to maximising the evidence lower bound (ELBO)~\cite{minka_divergence_2005}
\begin{equation}
KL(q(\theta_j^l)||p(\theta_j^l|x))= -(E_q[log p(\theta_j^l,x)]-E_q[log q (\theta_j^l)]
\end{equation}
where $E_q$ is the expected value under $q$. In practice the parameters are stochastic gradients sampled from the optimal variational distribution. 

Three variants of ELBO are evaluated in trained versions of the PM+MO framework. The first implementation (ELBOv1) samples $s$ from $q_i$ and computes the expected value in a basic form:
\begin{equation}
E_{q_{s(x)}}[log p(\theta_j^l,x)-log q_s(x)].
\end{equation}
ELBOv2~\cite{bingham_pyro_2018} uses the Rao-Blackwellization strategy proposed by Ranganath \textit{et al.} to reduce variance when estimating gradients by replacing random variables with conditional expectations of the variables~\cite{ranganath_black_2013}. The third variant ($\lambda KL$) also addresses variance in gradient estimation by including a term to limit the regularising influence of the KL term at initiation - and then scaling up the level as training progresses~\cite{sonderby_ladder_2016}. L1 and L2 norms in the training steps of variational inference present additional controls to regulate parameter updates. 

\section{Evaluation}
An analysis of the PM+MO framework comprises training and testing system variants on the task of multilabel classification on text and images from the MM-IMDb dataset. System performance is measured with $micro-F1$, $macro-F1$, $weighted-F1$, and $samples-F1$. $F1$ is a standard metric for measuring accuracy in multiclass classification tasks~\cite{madjarov_extensive_2012} and is computed as the mean of precision and recall
\begin{equation}
f_1^{sample} = \frac{1}{N}\sum_{N}^{i=1}\frac{2 |\hat{y_i}\cap y_i|}{|\hat{y_i}|+|y_i|} 
\end{equation}
where $N$ is the number of samples and $y_i$ is the tuple of predictions. Each of the four methods is an average of $F$-scores computed in the following ways: per sample (samples), across all system outputs (micro), by genre (macro), or by genre and with a weighted average on positive samples for each label.  Performance at system level is reported for all of these metrics and $weighted-F1$ is referenced in comparisons between systems in the text. 

\subsection{System Configuration}
Systems in the evaluation are all trained on a single Tesla K80 GPU and with a batch size of 512. Priors for the weights and biases of each layer in builds with variational inference are modeled using Laplace distributions initialised at $\mu=0.1$ and $\sigma=0.01$. Parameters for these distributions are learned during gradient steps with the three variants of ELBO detailed above. In the version with KL scaling ($\lambda KL$), the scaling term is set at $\lambda=0.2$ following tests in the range $(0.1,1.0)$. L1 and L2 updates to parameters are set by a different $\lambda= 0.1$ in all tests.

\subsection{Results}
Experiments aim to measure the impact of training with multiple objectives, variational inference, and probabilistic regularisation methods when conducting multimodal fusion for classification tasks. Assessment starts with a comparison of the best performing version of the PM+MO framework - PM+MO ($\lambda KL+L2$) -  with the GMU baseline. An ablation analysis on several PM+MO variants provides a granular analysis of regularisation methods associated with variational inference. Reported numbers are the means of scores calculated over five complete cycles of training and testing.

\begin{table}
\caption{F-scores for PM+MO and GMU Baseline (mean over 5 cycles)}\label{tab1}
\begin{tabular}{|p{4.00cm}||p{1.75cm}|p{1.75cm}|p{1.75cm}|p{1.75cm}|}
 \hline 
 {System} & \multicolumn{4}{c|}{F-score}\\
 \cline{2-5}
  & Micro & Macro & Weighted & Samples\\
 \hline
 PM+MO ($\lambda KL+L2$) & 0.620 & 0.549 & 0.617 & 0.620\\
 PM+MO ($\lambda KL+L2+1024$) & 0.602 & 0.524 & 0.599 & 0.607\\
 GMU Baseline   & 0.618 & 0.528 & 0.608 & 0.617\\
 \hline
\end{tabular}
\end{table}

Topline results for our proposed framework and the GMU baseline are presented in Table \hyperref[tab1]{1}. Hyperparameter settings were optimised for each system with differences in learning rate (PM+MO=0.005, GMU=0.001) and dropout (PM+MO=0.9, GMU=0.7). A version of the PM+MO framework with a fused embedding module including ELBO+KL scaling and L2 regularisation in the fused embedding model - and wide layers with 3000 neurons - scored higher on all $F$-scores than the GMU baseline ($weighted-F1 = +0.009$). The baseline combines a Gated Multimodal Unit and a simple classifier with maxout activation - and is trained end-to-end with a single binary cross-entropy objective. Regularisation and hyperparameter settings conform to specifications shared in the publication~\cite{arevalo_gated_2017} and repository. Code conversion from Theano to PyTorch is a contributor to the difference in scores for this build of the GMU system against those stated in the original research ($weighted-F1= 0.617$).

A version of our framework with 1024 neurons in each linear layer completes the topline analysis. Lower scores for this approach underline the benefits of training with wide layers when regularisation offsets variance. As a final check of the benefits of training with a combination of variational inference and wide layers, we tested a version of the GMU baseline with wide layers and observed lower accuracy to the reported system. Training time per epoch on a single GPU for the best performing PM+MO system is 5.25 secs compared to 1.10 secs for the GMU baseline. Total training time for the former is still low at 2m11s - but extended training times are significant considerations in large-scale data regimes~\cite{ma_democratizing_2018}. 
\newpage
\begin{figure}[H]
\includegraphics[width=1.0\textwidth]{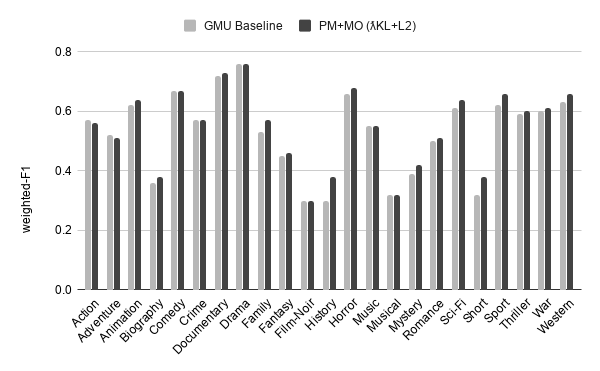}
\caption{PM+MO and GMU Baseline performance on genres by $weighted-F1$ (mean over 5 cycles).} \label{fig2}
\end{figure}

Measurements of accuracy on individual classes are presented in Figure \hyperref[fig2]{3}. The most performant PM+MO system reported higher $weighted-F1$ scores in relation to the GMU baseline for 15 of the 23 movie genres. Classification accuracy matched or exceeded the baseline on all genres where $weighted-F1$ for the latter system was less than 0.5.

\begin{table}
\caption{Ablation for PM+MO with F-scores (mean over 5 cycles)}\label{tab2}
\begin{tabular}{|p{4.00cm}||p{1.75cm}|p{1.75cm}|p{1.75cm}|p{1.75cm}|}
 \hline
 {Specification} & \multicolumn{4}{c|}{F-score}\\
 \cline{2-5}
  & Micro & Macro & Weighted & Samples\\
 \hline
PM+MO ($\lambda KL+L2$) & 0.620 & 0.549 & 0.617 & 0.620\\
PM+MO ($\lambda KL+L1$) & 0.612 & 0.545 & 0.613 & 0.611\\
PM+MO ($\lambda KL$) & 0.612 & 0.544 & 0.611 & 0.612\\
PM+MO (ELBOv2+L2) & 0.618 & 0.543 & 0.614 & 0.619\\
PM+MO (ELBOv2) & 0.615 & 0.541 & 0.611 & 0.616\\
PM+MO (ELBOv1+L2) & 0.617 & 0.544 & 0.614 & 0.618\\
PM+MO (ELBOv1) & 0.612 & 0.543 & 0.609 & 0.613\\
M+MO (2 units minus VI) & 0.611 & 0.529 & 0.602 & 0.611 \\
 \hline
\end{tabular}
\end{table}

Scores for several versions of the PM+MO framework are presented in Table \hyperref[tab2]{2} with the objective of comparing the impact of regularisation strategies. Hyperparameter settings are uniform across all runs with the exception of the specific methods noted in rows.  ELBO versions incorporating methods for managing variance outperform basic implementations of ELBO (ie ELBOv1). KL scaling with L2 regularisation delivers a marginal improvement ($weighted-F1 = +0.003$)  on the same configuration with Rao-Blackwellization (ELBOv2+L2). Supplementary L2 norm penalties on parameter updates boost accuracy on all configurations. A build with multiple objectives and excluding variational inference (M+MO) returns the lowest $weighted-F1$ ($-0.015$ w.r.t. PM+MO ($\lambda KL+L2$)).

\section{Related Work}
\subsection{Multimodal Representation Learning} 
Researchers have investigated the role of neural networks in combining representations from multiple modalities to perform end tasks for several decades~\cite{yuhas_integration_1989}. Multimodal fusion is deployed in classification tasks when all constituent modalities are present during training and inference~\cite{ngiam_multimodal_2011}. Coordinated embedding methods are an alternative method for these tasks~\cite{baltrusaitis_multimodal_2017}. Separation between vectors is retained in these approaches by projecting the textual and visual representations into a common $d-dimensional$ space and introducing a constraint~\cite{frome_devise_2013}. Fusion-based learning results in a single output vector: one advantage for sticking with this approach in our framework is to facilitate transfer between modules. Multimodal embeddings are also learned by Silberer and Lapata to perform word similarity and object classification~\cite{silberer_learning_2014}. The process proposed for this and related methods~\cite{silberer_learning_2014,bruni_distributional_2012} differs from the method in our system by including Singular Value Decomposition (SVD) to integrate constituent embeddings.

\subsection{Multiple Modules and Objectives} 
System architectures composed of multiple modules form a foundational area in the research on neural networks. A primary objective in this literature is the construction of classification frameworks that generalise to unseen samples~\cite{zhang_augmenting_2016}. Auda \textit{et al.}~\cite{auda_multimodal_1998} detailed several approaches to decomposing tasks and designed a classifier with multiple modules to solve components for separate sub-tasks. In this case, a voting layer acted as a constraint on outputs from individual components~\cite{auda_new_1994}. Early implementations of modular architectures for task decomposition were trained with a single objective - or were separated into distinct models. Secondary losses are implemented during training in the related areas of representation learning and transfer learning. Our system approximates Zhang et al’s proposal to introduce an auxiliary objective and module into a classification framework~\cite{zhang_augmenting_2016}. The method selection in this research differs to ours in implementing unsupervised learning for the auxiliary components. Du \textit{et al.}~\cite{du_adapting_2018} proposed a system for measuring cosine similarity between auxiliary and main losses when the former contributes to the latter~\cite{du_adapting_2018}. In contrast to our work, positive transfer with multiple losses are applied in instances where source and target tasks share related objectives. 

\subsection{Probabilistic Deep Learning} 
Probabilistic methods in this research extend an approach to machine learning where the assessment of architectures is based on inverse probability~\cite{mackay_bayesian_1992}. Here the plausibility of the model - or in our case, the parameters in each layer - are computed w.r.t. to the data. Variational inference is a non-deterministic method that replaces elements in probabilistic inference with approximations when the computation of integrals is intractable. A family of distributions is placed over the model parameters and the candidate distribution with the lowest KL divergence from the true posterior is selected~\cite{ranganath_black_2013}. ELBO formulates this minimisation as optimisation and rewards candidate distributions that maximise both $p(z|x)$ and a spread of uncertainty. The ELBO term in our system extends optimisation with simple operations to regularise parameter updates. Ma \textit{et al.}~\cite{ma_mae_2019} modify ELBO with a supplementary regularisation term to improve representation learning - although the objective of this technique is to reward diversity in the selection of candidate distributions~\cite{ma_mae_2019}. In contrast to our proposals, enhancements or substitutes for ELBO in this and other research are centred on variational autoencoder (VAE) architectures~\cite{alemi_fixing_2018,chen_infogan_2016}. The selection of Laplace distributions in our system is informed by the ability of these distributions to model data with a high level of heterogeneity~\cite{geraci_notebook_2018}. Stochastic variational inference and related methods are implemented in our system using the Pyro PPL~\cite{bingham_pyro_2018}.

\subsection{Regularisation Methods} 
Our framework retains means for reducing variance when conducting multimodal fusion proposed by Arevalo \textit{et al.}~\cite{arevalo_gated_2017}. Batch normalisation was introduced to minimise the impact of changes in parameters on the distributions for activations~\cite{ioffe_batch_2015}. Goodfellow \textit{et al.} describe the maxout activation function as an averaging technique in neural network-based architectures that compliments dropout~\cite{goodfellow_maxout_2013}. In initial testing, we verified the performance gain from selecting maxout activation in the classifier module and retained it in the PM+MO architecture. Contributions that describe interactions between parameters and optimiser algorithms~\cite{loshchilov_decoupled_2019,hanson_comparing_1989} supported our decision to test different forms of managing weight updates. The context differs as we extend these techniques to training representations with variational inference.

\section{Conclusion}
In this research, we have demonstrated that a framework of sub-modules trained with variational inference as one of multiple objectives leads to improvements in performance on multimodal classification. The proposed framework supports wide layers and higher learning rates when compared with systems trained with a single objective. Improvements in generalisation over multiple objective systems that exclude variational inference are also demonstrated. An evaluation of regularisation methods associated with variational inference underlines the advantages of probabilistic approaches in extending options to calibrate parameter updates during training and offset overfitting in multimodal systems. We plan to train on additional modalities and extend probabilistic methods in representation learning as further contributions to the research on improving generalisation in multimodal systems. 

\section*{Acknowledgements}\label{sec:acknowledgements}
The project leading to this publication has received funding from the European Union’s Horizon 2020 research and innovation programme under the Marie Skłodowska-Curie grant agreement No 812997.

%
%
%
\bibliographystyle{splncs03}

\end{document}